# WOLI at SemEval-2020 Task 12: Arabic Offensive Language Identification on Different Twitter Datasets

**Yasser Otiefy**
WideBot
y.otiefy@widebot.ai

**Ahmed Abdelmalek**
WideBot
a.sherif@widebot.ai

**Islam El Hosary**
WideBot
islam@widebot.ai

## Abstract

Communicating through social platforms has become one of the principal means of personal communications and interactions. Unfortunately, healthy communication is often interfered by offensive language that can have damaging effects on the users. A key to fight offensive language on social media is the existence of an automatic offensive language detection system. This paper presents the results and the main findings of SemEval-2020, Task 12 OffensEval Sub-task A Zampieri et al. (2020), on Identifying and categorising Offensive Language in Social Media. The task was based on the Arabic OffensEval dataset Mubarak et al. (2020). In this paper, we describe the system submitted by WideBot AI Lab for the shared task which ranked 10[th] out of 52 participants with Macro-F1 86.9% on the golden dataset under CodaLab username "yasserotiefy". We experimented with various models and the best model is a linear SVM in which we use a combination of both character and word n-grams. We also introduced a neural network approach that enhanced the predictive ability of our system that includes CNN, highway network, Bi-LSTM, and attention layers.

## 1 Introduction

Social media platforms such as Facebook and Twitter provide an online space for individuals or groups to express their opinions often with limited restrictions. Hence, social media users may use filthy, sexual, or offensive language Jay and Janschewitz (2008). Offensive phrases could ridicule or insult an individual or a group Razavi et al. (2010). Many applications involve offensive language detection, for instance, parental control or adult content filters; identifying hate speech that incites or presage violence Waseem and Hovy (2016); Identifying troll accounts Darwish et al. (2017); conflict detection, which is often lead up to verbal hostility Chadefaux (2014); quantifying the intensity of polarisation Conover et al. (2011). In addition to the huge effort required by human annotators, the risk of applying discrimination or favouritism should be placed under consideration. Moreover, a manual annotation task by human annotators would increase the system response times drastically, since a computer-based solution can perform this task much faster than humans. With the rapid growth in the amount of user-generated content in the social media services, the scalability of manual filtering highlights the need for automating the process of on-line hate-speech detection Pitsilis et al. (2018). OffensEval, SemEval-2020 Task 12 sub-task A Zampieri et al. (2020) aims to train a model on Arabic tweets collected from Twitter, annotated by human annotators to identify the offensive language within the messages. In this paper, we describe the system submitted by WideBot AI Lab for the shared task of offensive language identification in Arabic and discuss the results of our experiments and it is organised as follows: Section 2 discusses important details about the task and data used. It additionally demonstrates the previous work done on this task. Section 3 discusses all the datasets used in training. Section 4 presents an overview of the key algorithms. Section 5 explains the procedures of the experiments and evaluation metrics. Section 6 compares the results. Section 7 shows the conclusions and future works.

## 2 Background

The goal of this task is to distinguish between offensive and non-offensive messages. Offensive messages often contain insults, threats, and any form of targeted profanity. The dataset provided for the task is about 10,000 annotated messages. Each message is assigned one of the following two labels:

- Not Offensive (NOT): Posts that do not contain offence or profanity.

- Offensive (OFF): We label a post as offensive if it contains any form of non-acceptable language (profanity) or a targeted offence, which can be veiled or direct.

### 2.1 Related work

Previous works used a variety of classification techniques. For instance. Classical models used by (Kwok and Wang, 2013; Malmasi and Zampieri, 2017) and deep learning models (Agrawal and Awekar, 2018; Badjatiya et al., 2017; Nobata et al., 2016). Moreover, other works employed sentiment words and contextual features (Yin et al., 2009; Nobata et al., 2016) found that linguistic features such as tweet length, average word length, number of punctuation, number of discourse connectives can be useful for abusive language identification. Recently, In general, transfer learning achieved state-of-the-art performance in many natural language processing tasks and specifically in offensive language identification such as pre-trained word embeddings as a transfer learning by Badjatiya et al. (2017) and Bidirectional transformer for offensive language detection by Liu et al. (2019). Much research has been done on Arabic offensive language detection based on classical supervised algorithms. Abozinadah and Jones (2016) used SVM and novel ways of normalisation to deal with misspelling words. Later they addressed the problem using a statistical learning approach Abozinadah and Jones (2017). Also, Alakrot et al. (2018) proposed a solution with the SVM model with a combination of word-level features and n-gram features on a new dataset collected from YouTube. Deep learning approaches are widely used in offensive language detection, for example, Albadi et al. (2018) used a simple recurrent neural network (RNN) architecture with gated recurrent units (GRU) on pre-trained word embeddings. Moreover, Mubarak and Darwish (2019) introduced a deep learning model. More details can be found in this survey for a complete review Al-Hassan and Al-Dossari (2019).

## 3 Dataset

This subsection shows the used dataset beyond the provided training data by Arabic OffensEval shared task Mubarak et al. (2020) which are Twitter examples with labels NOT for non-offensive examples and OFF for offensive examples. We faced a dataset imbalance problem since the provided dataset has 1,915 examples for OFF and 8,085 examples for NOT, so we decided to use the following additional datasets in our experiment:

- **L-HSAB**: Levantine Arabic messages extracted from Twitter. The dataset contains 8,846 tweets with three labels: Hate, Abusive, and Normal. For further details and statistics, please refer to Mulki et al. (2019).

- **WideBot's offensive language dataset**: Our private dataset was generated by human annotators. They are messages written in the form of chat messages between two individuals. annotators include some properties and attributes in creating the dataset such as it should contain vulgar languages, phrases that try to harshly criticise or mock someone, and obscene and offensive terms. The dataset is balanced with 1600 offensive and 1600 non-offensive messages. A label of two labels was assigned to each message whether it is offensive (OFF) or not (NOT).

## 4 Proposed Models

This subsection shows the key algorithms and modeling decisions we used in our experiments. We used a Logistic Regression model on a bag of words of the training dataset as our baseline and other two main systems which are Linear SVM from classical models and the other one is based on recently developed neural network architectures.

## 4.1 SVM

The system used an SVM model Cortes and Vapnik (1995) with the linear kernel to train on an n-gram language model. For the feature representation, we used TF-IDF and combined the word n-grams with char n-grams. The feature matrix consists of both character (from uni-gram to 6-grams) and word (from uni-gram to 5-grams) levels.

## 4.2 Deep Learning Model

In this model, we use word embeddings as our text representation then we feed it into a deep model that uses different neural layers which are CNN, highway network, Bi-LSTM, and attention layers (CHBA). This model is inspired by (Kim et al., 2016; Srivastava et al., 2015; Liu and Lane, 2016). We incorporated both character and word embeddings as inputs to our network using FastText[1] which is a pre-trained skip-gram word embedding for the Arabic language and trainable character embedding with the random initialisation. Every input is processed in its branch. Character embeddings are processed using CNN layers by Kim et al. (2016) and Highway network by Srivastava et al. (2015). At the same time, the word embeddings are processed by passing it to a dropout layer and the output of both branches are concatenated. At this step, we are ready to pass it to our main network. This network is a bidirectional LSTM layer with an attention layer inspired by Liu and Lane (2016). The output from these layers is projected on a sequence of dense layers and finally the output layer which will map the output of these layers to one of the classes which are OFF or NOT. Figure 1 shows the architecture of this model.

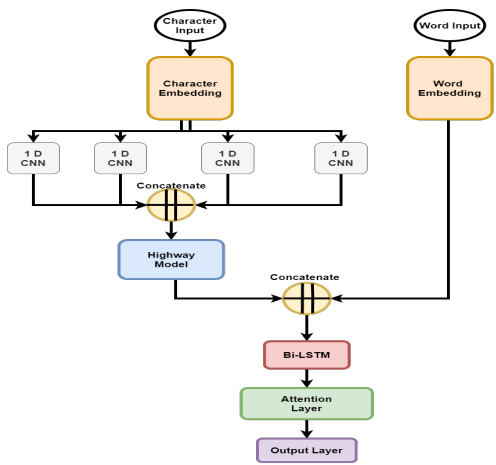

Figure 1: Generic architecture graph of the CHBA network

## 5 Experimental Setup

We train different models for the classification task. In this section, we explain the preprocessing and training phases. The classifiers we experimented with include SVM with features of a word and character n-grams and the CHBA network. We experimented using both the original dataset only and the external datasets mentioned before. We have chosen the SVM model with the OffensEval dataset and other external datasets since it performs the best score in our experiments until the time of submission, but we did further experiments using the CHBA network on the OffensEval dataset only with hyperparameter optimisation which did better than the submitted one. The evaluation metric of this task is Macro-F1 which calculates the unweighted-mean F1 of the two classes since the imbalance of the data classes affects the macro-F1, and usually, the score is penalised by the minority classes.

## 5.1 Preprocessing

Preprocessing is important in deriving useful information and improving the quality of data that directly affects the performance of our model, especially when dealing with text messages from various dialects and non-standard language. First, we did some preprocessing on each external dataset separately to make it fit in our preprocessing pipeline and these steps are:

- For L-HSAB dataset, we merge the two classes (Hate and Abusive) into one (OFF), and renamed the class "Normal" to be "NOT".

- For WideBot dataset, In an attempt to overcome the imbalanced classes problem in the original dataset, we decided to include only the offensive messages only.

---
[1] FastText: https://fasttext.cc/

Then we feed the dataset to a preprocessing pipeline. The pipeline contains the following:

- Removing all URLs, mentions, emails, dates, numbers, punctuation, English letters, stop words found in NLTK, and the emojis.

- Unifying similar Arabic characters such as Hamza and Yaa.

- Replace the consecutive repeated characters with only one character.

## 5.2 Data Split

We used the officially released dataset in addition to our datasets mentioned before and used the standard split (train/validation/test) with the standard dataset provided by the OffensEval task organisers. We extended the training set to be a subset of the standard train, L-HSAP, and WideBot's dataset. Eventually, we had 16,046 messages for the training set, 1,000 messages for the validation set, and 2,000 for the test set. All of our models were trained on the same data set.

## 5.3 SVM model

We added a special preprocessing for the SVM model, by converting the text into vectors of TF-IDF on both character and word levels. The vectors are ranging from uni-grams to 5 n-grams for word vectorizer and from uni-grams to 6 n-grams for character vectorizer. We tested the SVM with a linear kernel which achieved 87.3% on the validation set. Some feature reduction was done to include the features with the highest TF-IDF score on both character and word levels. In our experiment, we tried different combinations of features from both word and character TF-IDF vectorizer on the validation dataset to select the best combination. We start by removing the features with the least TF-IDF score and continue until we reach the best combination.

## 5.4 CHBA Network

We tried the skip-gram version of FastText Arabic embedding and Aravec by Soliman et al. (2017) with hidden layer size 300 as our word embedding in this experiment which is trained on about 67 million tweets, but FastText proved its effectiveness in this task. Each word is represented with a vector its size is 300 and each sentence is limited with a max sequence length 50 of words so the output should be (50) * 300, and for character embedding, we limited the word length with maximum 10 characters and character embedding size 20 with matrix shape (10) * 20. For activation layers we use $Tanh$ in the character processing branch and $Relu$ for the others, We set a drop-out rate by 0.5 for embedding layers and 0.33 for other layers. The dataset on hand is imbalanced, which means in the first few iterations the network is just learning the bias. To speed up the convergence, we set the bias on the logits such that the network predicts the probability of (positive/negative) at initialisation. Finally, for optimisation purposes, we used sparse categorical cross-entropy as our loss function and Adadelta as an optimiser as it converged very well in most of our experiments for this system. Overparameterized deep networks like CHBA, are prone to overfitting. With other regularisation methods, we used a technique called Flooding inspired by Ishida et al. (2020) to prevent the training loss from reaching zero and float around small constant b, it is named as flooding level by the original authors. It does not mean the training loss will become positive. We still can learn till zero training error, the purpose here is to only make the loss positive assuming the flooding level is not too big. Assume the original learning objective is $J$, the modified learning objective $\tilde{J}$ with flooding is

$$\tilde{J}(\theta) = |J(\theta) - b| + b$$

where *b* is the flooding level, a small positive constant, and $\theta$ is the model parameter. We experimented with multiple values for the hyperparameter *b*. When $J(\theta) > b$, the direction of J($\theta$) and $\tilde{J}$ w.r.t. $\theta$ will be the same, which means the learning objective is above the flooding level, in other words, there is a gravity effect with gradient descent. But when $J(\theta) < b$, flooding state, $\tilde{J}$ w.r.t. $\theta$ will point to the opposite direction of $J(\theta)$, which means there is a buoyancy effect with gradient ascent.

## 6 Results

Table 1 shows the Macro F1 score of all of our models experimented on task 12 OffensEval over the Arabic dataset. After the submission, we made further development in CHBA. It outperformed the SVM with a score of 88.72%, for more statistics about the scoreboard, you can find them in table 2.

| Classifiers | Dataset | Validation | Test | Notes |
|---|---|---|---|---|
| Baseline | Original data only | 75% | 76.59% | |
| Baseline | Adding external data | 81.19% | 83.66% | |
| SVM | Original data only | 84.83% | 85.61% | |
| SVM | Adding external data | 87.32% | 86.91% | The submitted model |
| CHBA | Original data only | 88.54% | 88.13% | |
| CHBA | Adding external data | **88.77%** | **88.72%** | |

Table 1: Macro F1 score of different models

The confusion matrix in table 3 shows in detail the error pattern of our classifier, which is better in detecting the not offensive text messages than detecting the offensive ones, we think this is because of the challenging of the subtle meanings and the context which are too hard to be caught by the model.

| Mean | Max | Min | Standard Deviation |
|---|---|---|---|
| 79.24% | 90.17% | 44.51% | 12.19% |

Table 2: Statistics about Macro F1-Score in OffensEval scoreboard

| | Actual NOT | Actual OFF |
|---|---|---|
| Predicted NOT | 1,540 | 103 |
| Predicted OFF | 58 | 299 |

Table 3: Confusion matrix of the winning model (Linear SVM)

**Note:** All the evaluation was done on the golden test set and validation set provided by OffensEval Team.

## 7 Conclusion

In this paper, we presented the results and the main findings of SemEval-2020 Task 12 OffensEval - Arabic Task on Identifying and categorising Offensive Language in Social Media in which we carried out different experiments with Linear SVM using different features and CHBA network. We also used a feature selection approach to deal with the high sparsity of the data, also we have shown how stacking the TF-IDF on word n-grams and character n-grams of the task dataset with external datasets could improve the performance of the model. However, hyperparameter optimisation and the usage of pre-trained word embeddings for the CHBA network showed its effectiveness in this task as it outperformed the linear SVM model knowing that we just used the dataset provided by OffensEval only. Future work includes investigating text representations, improving the deep learning model, and releasing WideBot conversational dataset from different dialects for offensive language identification tasks.